%% file: main.tex
\documentclass[sigconf]{acmart}
\usepackage[ruled,vlined]{algorithm2e}
\usepackage{subcaption}
\usepackage{multicol}
\usepackage{multirow}
\usepackage[utf8]{inputenc}
\usepackage{graphicx}
\usepackage{booktabs}
\usepackage{amsmath}
\usepackage{mathrsfs}
\AtBeginDocument{%
  \providecommand\BibTeX{{%
    \normalfont B\kern-0.5em{\scshape i\kern-0.25em b}\kern-0.8em\TeX}}}

\begin{document}

\title{Like Article, Like Audience: Enforcing Multimodal Correlations for Disinformation Detection}

\author{Liesbeth Allein}
\affiliation{%
  \institution{European Commission, Joint Research Centre (JRC)}
  \city{Ispra}
  \state{Lombardy}
  \country{Italy}
}

\author{Marie-Francine Moens}
\affiliation{%
  \institution{KU Leuven}
  \city{Leuven}
  \country{Belgium}
}

\author{Domenico Perrotta}
\affiliation{%
  \institution{European Commission, Joint Research Centre (JRC)}
  \city{Ispra}
  \state{Lombardy}
  \country{Italy}
}

\renewcommand{\shortauthors}{Allein et al.}

\begin{abstract}
  User-generated content (e.g., tweets and profile descriptions) and shared content between users (e.g., news articles) reflect a user's online identity. This paper investigates whether correlations between user-generated and user-shared content can be leveraged for detecting disinformation in online news articles. We develop a multimodal learning algorithm for disinformation detection. The latent representations of news articles and user-generated content allow that during training the model is guided by the profile of users who prefer content similar to the news article that is evaluated, and this effect is reinforced if that content is shared among different users. By only leveraging user information during model optimization, the model does not rely on user profiling when predicting an article's veracity. The algorithm is successfully applied to three widely used neural classifiers, and results are obtained on different datasets. Visualization techniques show that the proposed model learns feature representations of unseen news articles that better discriminate between fake and real news texts.
\end{abstract}

\ccsdesc[300]{Computer systems organization~Neural networks}
\ccsdesc[300]{Theory of computation~Semi-supervised learning}
\ccsdesc[500]{Information systems~Document representation}
\ccsdesc[100]{Information systems~Clustering and classification}
\ccsdesc[300]{Computing methodologies~Natural language processing}
\ccsdesc[500]{Computing methodologies~Learning latent representations}

\keywords{Disinformation, Fake News, Constrained Representation Learning, Multimodal Data Fusion, Natural Language Processing}


\maketitle

\input{introduction}
\input{related_work}
\input{methodology}
\input{experiments}
\input{discussion}
\input{conclusion}

\bibliographystyle{ACM-Reference-Format}
\bibliography{acmart}

\appendix

\end{document}

%% file: introduction.tex
\section{Introduction}

Disinformation - or the more politically-loaded term `fake news' - is not new. Throughout history, it has been a popular means of deception and persuasion. Although the motivations for disinformation have remained fairly unchanged, the Internet and social media nowadays provide low-resource platforms to disseminate deceitful and inaccurate information more rapidly and broadly than ever before. 
The authors who create \textit{articles containing disinformation} and the \textit{social media users} who consequently spread them are key components of disinformation dissemination. This paper focuses on the (cor)relations between online news articles and social media users, and examines whether computational disinformation detection models benefit from leveraging such correlations.

\begin{figure}
    \centering
    \includegraphics[width=7cm]{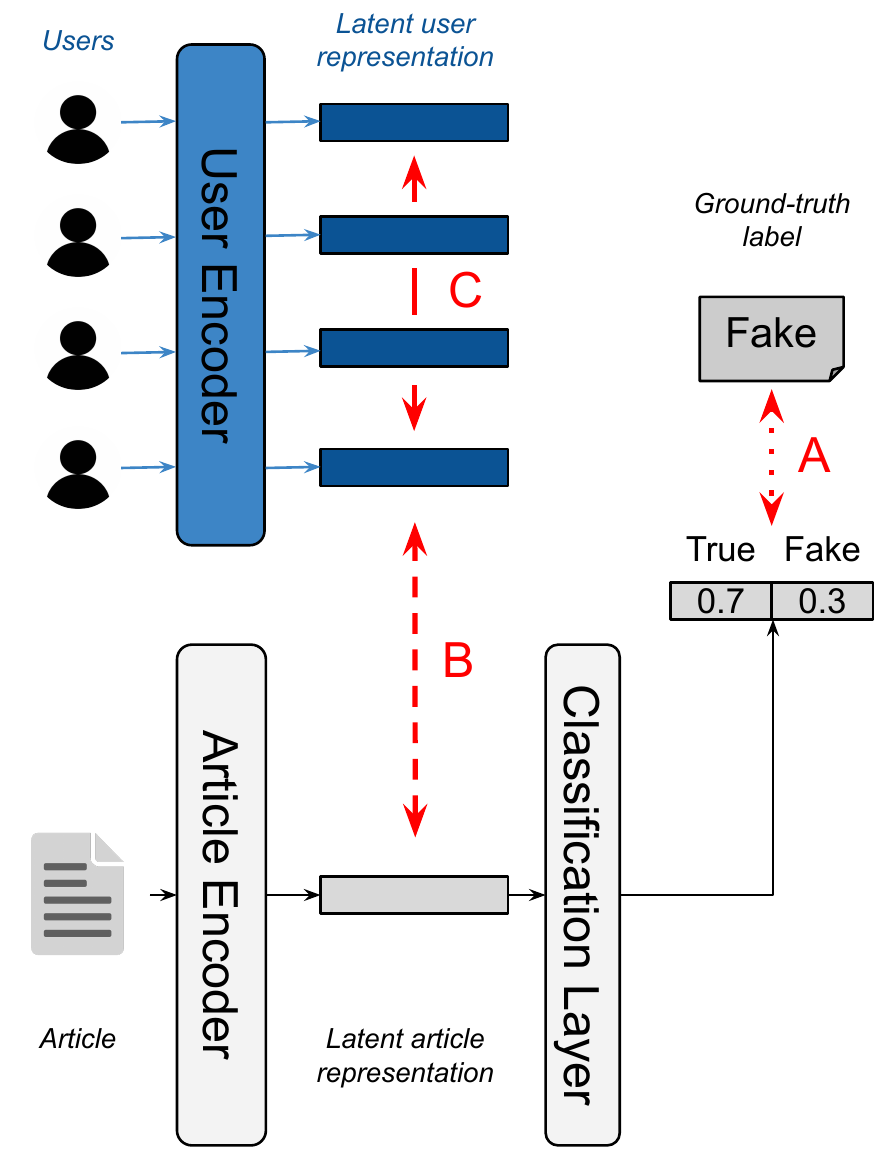}
    \caption{Overview of the multimodal disinformation detection algorithm. Given an article, a classifier encodes the article and predicts whether it is `true' or `fake' (in grey). A subset of users who shared the article is encoded independently (in blue). All parameters are optimized on three training objectives (in red): discriminate between true and fake news (A), minimize the cosine distance between the article latent representation and the user latent representations (B), and minimize the cosine distance between the user latent representations (C).}
    \label{fig:overview_algorithm}
\end{figure}

The authors' and users' motivations for spreading information are twofold: \textit{audience-oriented} and \textit{self-oriented}. Firstly, they both target an ideal audience when presenting information and aim to elicit certain reactions from them (\textit{audience-oriented motivations}). In case of disinformation, authors aim to deceive and/or persuade their readers. Such audience-oriented motivations are often reflected in the content and style of disinformation articles: controversial topics, emotion-provoking words and persuasion rhetorics \cite{shu2017fake, przybyla2020capturing}. Users sharing those articles are not necessarily driven by the same audience-oriented motivations as the authors. They might want to simply inform their friends, entertain their followers or even debunk the information presented in the disinformation article \cite{giachanou2020role}. Despite their possibly contradicting audience-oriented motivations, both authors and users aim to construct an online social identity through the information they create and share (\textit{self-oriented motivations}). The information a person generates or shares is said to reflect the meta-image of the self \cite{marwick2011tweet}. Considering Twitter, a user's online identity is arguably contained in both their tweets (\textit{user-generated content}) and the external content they share (\textit{user-shared content}). Assuming that a user does not display a plethora of online identities within a single user profile, \textit{a user's online identity portrayed in their user-generated information (e.g., tweets and profile description) should somehow correlate with the identity reflected in the user-shared article}.

In this paper, we leverage correlations between and within article and user modalities for detecting disinformation. We construct a learning algorithm in which model parameters are optimized on three training objectives: (1) discriminate between true and false news articles, (2) minimize the distance between article and user latent representations, and (3) minimize the distance between latent representations of users who share the same article (Figure \ref{fig:overview_algorithm}). We apply this learning algorithm to three popular neural text classifiers used for disinformation detection: CNN \cite{kim-2014-convolutional}, HAN \cite{yang2016hierarchical} and DistilBERT \cite{sanh2019distilbert}. We train and evaluate their performance using the FakeNewsNet \cite{shu2020fakenewsnet} and ReCOVery \cite{zhou2020recovery} dataset. Our main contributions can be summarized as follows:
\begin{enumerate}
    \setlength{\itemindent}{0em}
    \item We design a multimodal learning algorithm for detecting online disinformation based on observations of human online behaviour and identity made in social sciences.
    \item In the algorithm, we propose to explicitly model unimodal and multimodal correlations by enforcing distance constraints between article and user latent representations during model training.
    \item Coordinating the article latent space using a loss function that incorporates learned user representations is - to our knowledge - new in multimodal disinformation detection.
    \item In contrast to existing multimodal disinformation detection models, our algorithm only leverages user information during model optimization. The prediction model does not rely on user profiling when predicting an article's veracity. This way, we respect the European Commission's ethics guidelines for trustworthy AI \cite{ethicsguidelines}.
    \item Statistical visualization techniques prove that our multimodal learning algorithm forces the models to learn feature representations that are better for discriminating between fake and true news.
\end{enumerate}

%% file: related_work.tex
\section{Related Work}
In this section, we discuss how previous studies have represented and leveraged user information for online disinformation detection.

A few disinformation detection tasks purely focus on the users. Those tasks aim to predict whether a given user is a fake news spreader \cite{rangel2020overview}, or try to distinguish between humans and bots \cite{ferrara2016rise}. In these tasks, users are mainly represented by predefined personality traits and linguistic properties that are extracted from tweets, often in combination with supporting metadata (e.g., number of statuses and followers) \cite{giachanou2021detection,balestrucci2020credulous}.

Disinformation detection models predicting the veracity of a given text (e.g., news article or tweet) sometimes take users as additional input. Here, users are represented in terms of their interaction with the text and/or their connection with other users in the dataset. 
\citet{kim2018leveraging}, for example, represent each user by a designated numerical identifier (ID) and construct (user, timestamp, article)-tuples. This way, a model knows not only who interacted with the articles, but also when and in which sequence the interactions occurred. \citet{ruchansky2017csi} construct similar tuples, but instead of merely representing users by their ID, they compute a binary incidence matrix of articles with which the users have engaged before. This way, users who engaged with the same articles will have similar representations. However, such approaches lack rich user representations and merely model users by their interaction with an article. Moreover, interactions between users remain rather implicit.

Some works include rich user information and describe the type of user-article interaction. Users are represented by user-generated texts such as their comment/reply to a news article or tweet \cite{shu2019defend,qian2018neural,zhang2019reply} and the retweet in which they comment on the original tweet \cite{song2019ced}. Although such user representations contain a user's stance and opinion towards an article, users are now merely represented by a single short text that is related to the article in question. This contrast to our approach as we represent a user by a larger collection of user-generated texts (i.e., profile description and tweets) that are not exclusively linked to a single article.

Finally, other works explicitly model users, user-article interactions and user-user interactions. They often do this by constructing a heterogenous graph in which user nodes are linked to both article nodes and other user nodes in various manners. \citet{nguyen2020fang}, for example, represent users by a semantic vector representation of their Twitter profile description. They then compute a social context graph in which user nodes are connected with articles nodes using stance edges and other user nodes denoting followership. \citet{chandra2020graph} represent users by the BOW vector over all the articles they have shared. The user-article edges simply denote that the user has shared the article while user-user edges show that the users have a follower-following relation. Although social groups can be arguably deducted from a user’s explicit social connections, we refrain from modeling such user-user relations as like-minded people are not necessarily connected to each other on social media. Moreover, explicit social network connections do not always mean that the two connected users share common interests: they can merely denote some social relation such as kinship \cite{fani2017temporally}. Unlike other popular social media platforms, the reciprocity level on Twitter is fairly low: less than one out of four social relations are mutual and about one out of three users do not have a reciprocal relation with any of the Twitter accounts they follow \cite{kwak2010twitter}. This suggests that people more likely use Twitter as a source of information than a social networking site \cite{kwak2010twitter}. We therefore opt to model user-user relations in terms of the common articles they have shared online, and look for correlations in their generated content.


%% file: methodology.tex
\section{Methodology}
We approach disinformation detection as a binary classification task. Given an online news article $a_i \in A$, a disinformation detection model transforms $a_i$ to its latent representation $h(a_i)$ and predicts whether $a_i$ is `fake/unreliable' ($y_i = 0$) or `true/reliable' ($y_i = 1$). In this paper, we leverage user information from Twitter users in the optimization step. Given a subset of users $U_i \subseteq U$ who shared article $a_i$ on Twitter, a separate encoder transforms each user $u_j \in U_i$ to their latent representation $h(u_j)$. During model training, the model is optimized on three objectives: 
\begin{enumerate}
    \item Minimize the cross-entropy between the predicted and ground-truth label probabilities for $y_i$ (= model output).
    \item Minimize the mean cosine distance between the article representation $h(a_i)$ and all user latent representations $h(u_j) \in U_i$.
    \item Minimize the mean cosine distance between all user latent representations $h(u_j) \in U_i$.
\end{enumerate}

The three training objectives are combined in a single loss function as a weighted sum. Both the disinformation detection model and the user encoder are optimized using the combined loss. As user information is only leveraged in the loss function during model training, the disinformation detection model only relies on $a_i$ for predicting $y_i$ (see Figure \ref{fig:overview_algorithm}).

In this section, we first discuss the three neural text classifiers to which we will apply our multimodal learning algorithm. We then elaborate on the various training objectives and explain how they are integrated in the learning algorithm.

\subsection{Model Architecture}

The learning algorithm takes the following features as input:
\begin{enumerate}
    \item $A = \{(a_i,y_i) |0<i\leq N\}$ is the set of $N$ \textbf{news articles}, where each article $a_i$ is represented by its title $t_i$ and body text $b_i$, and labeled with a ground-truth veracity label $y_i \in \{0,1\}$ (0 = `fake/unreliable', 1 = `true/reliable'). Each token in title $t_i$ and body text $b_i$ is represented by a one-hot vector that refers to a unique entry in either the GloVe vocabulary \cite{pennington2014glove} (CNN, HAN) or DistilBERT vocabulary \cite{devlin2019bert} (DistilBERT)\footnote{CNN/HAN: We use the standard tokenizer from the NLTK toolkit \cite{bird2009natural} and the GloVe 6B vocabulary of 400k unique tokens with 300-dimensional word embedding pretrained on Wikipedia 2014 and Gigaword 5 (uncased, 6 billion tokens). DistilBERT: We adopt the pretrained DistilBERT model and tokenizer (`distilbert-base-uncased') from the Huggingface Transformers library \cite{wolf-etal-2020-transformers}, with a vocabulary of 30,522 unique tokens and 768-dimensional pretrained word embeddings.}. Ultimately, $a_i$ is a concatenation of $t_i$ and $b_i$: $a_i = [t_i;b_i]$.
    \item $U = \{u_j|0<j\leq K\}$ is the set of $K$ \textbf{Twitter users} that shared at least one $a_i \in A$. Each $u_j$ is represented by their profile description $d_j$ and user timeline ${tw}_j$. Each token in description $d_j$ is represented by a one-hot vector that refers to a unique entry in either the GloVe vocabulary (CNN, HAN) or DistilBERT vocabulary (DistilBERT). Analogous to $d_j$, the tokens in a tweet in a user's timeline $tw_j$ are represented by one-hot vectors. User timeline ${tw}_j$ is a concatenation of the tweet vectors. Ultimately, $d_j$ and ${tw}_j$ are concatenated to get user representation $u_j$:  $u_j = [d_j;{tw}_j]$.
    \item $U_i \subseteq U$ is the \textbf{subset of users who shared the same article} $a_i$. User subset $U_i$ is represented as a $S{\times}V$ matrix, with $S$ the number of users who have shared $a_i$ and $V$ the number of vector values in $u_j$. For sake of memory usage, we set $S = 10$.\footnote{We found that increasing $S$ has little effect on model performance.} The $S$ users for each $U_i$ are automatically obtained by taking the user IDs linked to the $S$ lowest tweet IDs in an article's tweet ID list. If a user has shared multiple articles in the dataset, $u_j$ can belong to multiple user subsets.
\end{enumerate}

News articles $A$ and Twitter users $U$ are encoded independently but projected onto the same latent space, i.e., their latent representations have the same vector dimensions. In the multimodal learning algorithm, $A$ is used as input to the disinformation detection model while $U$ is leveraged in the loss function for model optimization.

\subsubsection{Disinformation Detection Model: Article Encoding and Classification} 

The disinformation detection model consists of an article encoder and a single linear classification layer.
The article encoding layer takes as input article $a_i$ and transforms it to a single latent vector representation $h(a_i)$:
\begin{equation}
    h(a_i) = encoding(a_i)
\end{equation}
We experiment with several text encoding methods that transform a given text to a single document representation: 
\begin{itemize}
    \item \textbf{CNN} \cite{kim-2014-convolutional}: The Convolutional Neural Network (CNN) performs one-level encoding: word-level encoding using three convolutional layers with a max-pooling operation. For this model, the length of $a_i$ is limited to 500 tokens. 
    The word embedding layer, initialized with pretrained GloVe embeddings, first transforms each token in $a_i$ to its 300-dimensional word embedding. These are then fed to three separate convolutional layers with different filter windows (3, 4, 5) with 100 feature maps each and ReLU activation. Each layer yields a 100-dimensional hidden article representation which are ultimately concatenated into a single 300-dimensional latent article representation $h(a_i)$, as done in \cite{kim-2014-convolutional}.
    \item \textbf{HAN} \cite{yang2016hierarchical}: The Hierarchical Attention Network (HAN) performs two-level encoding: word-level and sentence-level encoding using bidirectional GRU networks with attention mechanism. To perform two-level encoding, $a_i$ is transformed from a one-dimensional vector to a two-dimensional $Z{\times}W$ matrix with $Z$ the maximum number of sentences\footnote{We consider title $t_i$ as a single sentence, while body text $b_i$ is split into sentences using the sentence tokenizer from the NLTK toolkit \cite{bird2009natural}. For user encoding (Subsection \ref{user_encoding}), $d_j$ and each tweet in $tw_j$ are considered as single sentences.} (50) and $W$ the maximum number of tokens per sentence (50). 
    The word embedding layer, initialized with pretrained GloVe embeddings, first transforms each token in $a_i$ to its 300-dimensional word embedding. A bidirectional GRU with word-level attention and hidden size (50) then computes $Z$ latent sentence representations, where each sentence representation is the concatenated 100-dimensional last hidden state of the bidirectional GRU. Ultimately, a bidirectional GRU with sentence-level attention and hidden size (50) yields a single 100-dimensional latent article representation $h(a_i)$ (as done in \cite{yang2016hierarchical}), which is the concatenated last hidden state of the bidirectional GRU.
    \item \textbf{DistilBERT} \cite{sanh2019distilbert}: DistilBERT is a light Transformer model based on the BERT architecture \cite{devlin2019bert}. For this model, the length of $a_i$ is limited to 512 (= maximum input dimension of DistilBERT). The embedding layer transforms each token in $a_i$ to its 768-dimensional word embedding. These are then sent through five Transformer blocks with multihead self-attention. The last hidden states are max-pooled, resulting in a 768-dimensional latent article representation $h(a_i)$, as done in \cite{sanh2019distilbert}.
\end{itemize}

The latent article representation $h(a_i)$ is ultimately sent through a single linear layer with softmax activation:
\begin{equation}
    p_i = softmax(linear(h(a_i)))
\end{equation}
with $p_i$ the probability distribution over all classification labels.

\subsubsection{User Encoding}\label{user_encoding}
The user encoder takes as input user subset $U_i$ and transforms each $u_j \in U_i$ to its latent user representation $h(u_j)$: 
\begin{equation}
    h(u_j) = encoding(u_j)
\end{equation}
We use the same preprocessing steps and text encoding methods for $u_j$ as for $a_i$. During training, we adopt the same encoding architecture for both the article and user encoder: CNN article encoder + CNN user encoder, HAN article encoder + HAN user encoder, DistilBERT article encoder + DistilBERT user encoder. The resulting latent user representations $h(u_j)$ are combined in a $S{\times}L$ matrix:
\begin{equation}
    h(U_i) = [h(u_1);...;h(u_j)]
\end{equation}
with $S$ the number of users in $U_i$ and $L$ the dimension of $h(u_j)$: $L$ = 300 (CNN), $L$ = 100 (HAN), $L$ = 768 (DistilBERT). The latent user representation matrix $h(U_i)$ is eventually used as input to the loss function.

\subsection{Training Objectives}

\begin{figure}
    \centering
    \includegraphics[width=\linewidth]{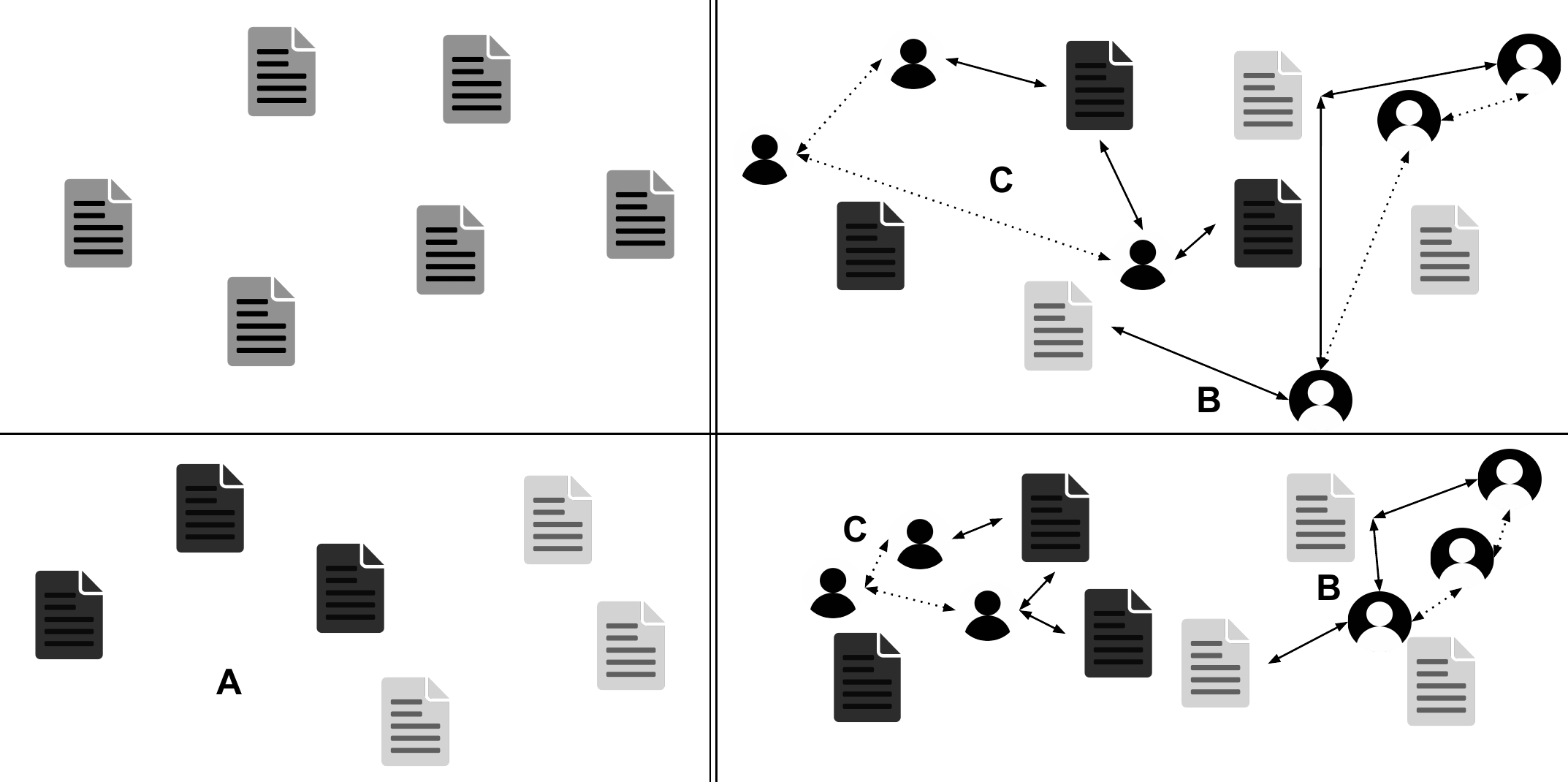}
    \caption{Overview of the training objectives. Given a set of articles (\textit{top left}), a disinformation detection model should be able to discriminate between true and fake articles (Objective A, \textit{bottom left}); this is a common classification objective. In our learning algorithm, the model is additionally forced to project not only the article and the users who shared it (Objective B), but also the users who shared the same article (Objective C) close together in the latent space (\textit{top right}$\xrightarrow{}$\textit{bottom right}).
    }
    \label{fig:constraints}
\end{figure}

In the multimodal learning algorithm, both the disinformation detection model and user encoder parameters are optimized on the following objectives (Figure \ref{fig:constraints}):
\begin{itemize}
    \item[(A)] \textit{The model should be able to discriminate between true and fake news articles.}
    \item[(B)] \textit{The model should project articles and the users who shared them closely to each other in the latent space.}
    \item[(C)] \textit{The model should project the users who shared the same article closely to each other in the latent space.}
\end{itemize}

We assign a dedicated loss function for each training objective and combine the losses in a single loss function as a weighted sum.

\paragraph{Objective A: Discriminate between true and fake articles}
A discriminative loss on the ground-truth and predicted labels is commonly used to optimize a classifier's parameters. We opt for the cross-entropy loss on the veracity prediction output, which is the probability distribution over the two labels yielded by a softmax function (= model output):
\begin{equation}
    \mathcal{L}_{pred} = - \sum\limits_{i}y_i\log\hat{y}_i 
\end{equation}
where \(y_i \in \{0,1\}\) is the ground-truth label for article $a_i$, and \(\hat{y}_i\) is the predicted probability for \(y_i\). By minimizing this \textbf{prediction loss}, the model is encouraged to look for patterns that discriminate between true and false articles, while it simultaneously learns patterns within each class. We provide baselines that are solely optimized on this prediction loss (\textit{base}).

\paragraph{Objective B: Find correlations between an article and the users who shared it}
We minimize the distance between an article's latent representation $h(a_i)$ and the latent representations of the users who shared that article $h(U_i)$. We use the cosine distance, as computed in \citet{kocher2017distance}, as distance measure. The \textbf{article-user distance loss} computes the arithmetic mean over the cosine distance between $h(a_i)$ and each $h(u_j)$ element of $h(U_i)$. If $a_i$ does not have a user subset $U_i$, or $U_i = \emptyset$, the article-user distance loss is set to $0$.      
\begin{equation}
    \mathcal{L}_{dist(a, U)} = 
    \begin{cases}
    \frac{1}{|U_i|}\sum\limits_{j=1}^{|U_i|}dist_{cosine}(h(a_i),h(u_j)) & \text{if } U_i \neq \emptyset \\
    0 & \text{otherwise}
    \end{cases}
\end{equation}

where $L$ is number of vector values for $h(a_i)$ and $h(u_j)$. By optimizing the model on the article-user distance loss, the model learns correlations between the article and its audience. 

\paragraph{Objective C: Find correlations between users who shared the same article}
We minimize the distance between the latent representations of all users $h(u_j)$ within the same user subset $h(U_i)$. Given user representation matrix $h(U_i)$, the \textbf{user-user distance loss} first computes a distance matrix $\mathbf{D}$ where $d_{ij}$ is the cosine distance between $h(u_i)$ and $h(u_j)$. 
\begin{equation}
    \mathbf{D} =
    \begin{bmatrix}
     d_{11} & d_{12} & \cdots & d_{1S} \\
     d_{21} & d_{22} & \cdots & d_{2S} \\
     \vdots & \vdots & \ddots & \vdots\\
     d_{S1} & d_{S2} & \cdots & d_{SS} \\
    \end{bmatrix}
    = (d_{ij}) \in \mathbb{R}^{S \times S}
\end{equation}
with $S$ the length of user subset $U_i$. Then, the average of each matrix column is calculated - leaving out a user's distance to itself ($ = d_{ii}$). This results in distance vector $d = [d_1, d_2,..., d_S]$, in which $d_i$ denotes the average cosine distance between $h(u_i)$ and all other $h(u_j)$ element of $h(U_i)$. Finally, the user-user distance loss is obtained by computing the arithmetic mean over $d_i$. If $a_i$ does not have a user subset $U_i$, or $U_i = \emptyset$, the user-user distance loss is set to $0$.
\begin{equation}
    \mathcal{L}_{dist(U)} = 
    \begin{cases}
    \frac{1}{|U_i|}\sum\limits_{j=1}^{|U_i|}dist(h(u_j), h(U_i)) & \text{if } U \neq \emptyset \\
    0 & \text{otherwise}
    \end{cases}
\end{equation}
\begin{equation}
    dist(h(u_j), h(U_i)) = \frac{1}{|U_i|-1}\sum\limits_{k=1}^{|U_i|}dist_{cosine}(h(u_j),h(u_k))
\end{equation}
By optimizing the model on the user-user distance loss, the model learns correlations between the users who shared the same news article.

\paragraph{Combined loss}
The combined loss is a weighted sum over the prediction loss $\mathcal{L}_{pred}$, the article-user distance loss $\mathcal{L}_{dist(a,U)}$ and the user-user distance loss $\mathcal{L}_{dist(U)}$:
\begin{equation}
    \mathcal{L} = \lambda_1\mathcal{L}_{pred} + \lambda_2\mathcal{L}_{dist(a,U)} + \lambda_3\mathcal{L}_{dist(U)}
\end{equation}
where $\lambda_1$, $\lambda_2$ and $\lambda_3$ sum to 1. During training, the data is fed to the model in batches, and the disinformation detection model and user encoder parameters are optimized using the mean batch loss after each forward pass.

 

%% file: experiments.tex
\section{Experiments}\label{experiments}
\subsection{Data}
\paragraph{FakeNewsNet} \cite{shu2020fakenewsnet}. The FakeNewsNet dataset comprises news articles from two fact-check domains: \textit{Politifact} and \textit{Gossipcop}. The articles are labeled as `fake' (0) or `true' (1). For each article, the dataset provides a URL to the original article and a list of tweet IDs that shared that specific article. We use the download script provided by the authors to crawl the articles (title and body text), tweet IDs (metadata) and user timelines (content and metadata of max. 200 tweets). In total, 568 \textit{Politifact} and 16,963 \textit{GossipCop} complete articles could be automatically crawled. To obtain the user subsets $U_i$ for each article $a_i$, we extract the user IDs from the tweets' metadata in the dedicated tweet ID list, link each user ID with its user timeline, extract the timeline's tweet content ($tw_j$) and take the user profile description from the metadata of the most recent tweet in the user timeline ($d_j$). As we can extract for each article the complete user information of at least one Twitter user, there are no empty user subsets $U_i$ for $Politifact$ and $GossipCop$ articles during training.  We regard \textit{Politifact} and \textit{Gossipcop} as two distinct datasets, as they greatly differ in scope: \textit{Politifact} focuses on political news while \textit{GossipCop} mainly verifies gossip news. 
\paragraph{ReCOVery} \cite{zhou2020recovery}. The \textit{ReCOVery} dataset contains 2,029 news articles about COVID-19 that are labeled as `unreliable' (0) or `reliable' (1). As these labels are similar to the `fake' (0) and `true' (1) labels in the other two datasets, we consider them synonymous. The dataset provides the full articles (title and body text) and a list of tweet IDs for each article. 
Given the tweet ID lists, we use the Twitter API to crawl the tweets' metadata, the user IDs and the user timelines (content and metadata). We obtain the user subset $U_i$ for each article $a_i$ in the same way as done for \textit{Politifact} and \textit{GossipCop}. However, only 133 articles have non-empty user subsets. 

\begin{table}[]
    \small
    \centering
    \begin{tabular}{cccc}
    \toprule
        Domain & True/Reliable & Fake/Unreliable & Total \\ 
    \midrule
        Politifact & 248 & 320 & 568 \\ 
        Gossipcop & 12,904 & 4,059 & 16,963 \\ 
        ReCOVery & 1,364 & 665 & 2,029 \\ 
    \midrule
        \textit{Total} & 14,516 & 5,044 & 19,560 \\
    \bottomrule
    \end{tabular}
    \caption{Overview of the three datasets.}
    \label{tab:datasets}
\end{table}

\subsection{Experimental Setup}

As the Politifact and ReCOVery dataset do not contain a high number of data samples, we combine the three datasets in one dataset. We split the combined dataset in a train (80\%), validation (10\%) and test (10\%) set in a label-stratified manner (random seed = 42). 
Although we train and validate on the combined dataset, we report performance results for each dataset individually. The results are obtained in terms of recall, precision and F1. During model training, the data is fed in batches of 32 (CNN/HAN) or 8 (DistilBERT). We investigate four experimental setups:
\begin{enumerate}
    \item \textit{base}: User information is not leveraged. The model parameters are optimized using only the prediction loss $\mathcal{L}_{pred}$.
    \item \textit{+u/d}: Users are represented by their profile description ($d_j$). Both the model and user encoder parameters are optimized using $\mathcal{L}$, which is the weighted sum of $\mathcal{L}_{pred}$, $\mathcal{L}_{dist(a,U)}$ and $\mathcal{L}_{dist(U)}$.
    \item \textit{+u/t}: Users are represented by their tweets ($tw_j$). Both model and user encoder parameters are optimized using $\mathcal{L}$.
    \item \textit{+u/d+t}: Users are represented by both their profile description ($d_j$) and tweets ($tw_j$). Both the model and user encoder parameters are optimized using $\mathcal{L}$.
\end{enumerate}
The Adam optimization algorithm (learning rate = 1e-4) optimizes the parameters after every forward pass. We perform early stopping on the validation loss with patience (7). 

To decide the $\lambda_1$, $\lambda_2$ and $\lambda_3$-values in $\mathcal{L}$ for each model, we experiment with different value combinations for CNN$_{+u/d}$, HAN$_{+u/d}$ and DistilBERT$_{+u/d}$. We additionally tested whether or not including both $\mathcal{L}_{dist(a,U)}$ and $\mathcal{L}_{dist(U)}$ yields higher model performance than simply including one of the two losses by setting either $\lambda_2$ or $\lambda_3$ to zero. Based on the validation set, we found the following optimal [$\lambda_1$, $\lambda_2$, $\lambda_3$]-values: [0.8, 0.1, 0.1] (CNN), [0.5, 0.25, 0.25] (HAN), [0.33, 0.33, 0.33] (DistilBERT). This shows that both correlation losses contribute to

\subsection{Results}

Table \ref{tab:overall_results_fake} and Table \ref{tab:overall_results_true} show the performance results in terms of precision (P), recall (R) and F1-score (F1) for the fake/unreliable and true/reliable label, respectively. We opt for these performance metrics as the two classes are rather unbalanced.
The results show that leveraging user information in a multimodal learning algorithm positively influences the model performance for all three neural classifiers and datasets. 

For the fake class (Table \ref{tab:overall_results_fake}), the user setup with only user profile descriptions ($+u/d$) has the highest positive impact when predicting Politifact articles: +3.78/+9.09/+9.19\% (P/R/F1) for the CNN model, and +6.02/(-3.03)/+0.35\% (P/R/F1) for the DistilBERT model.
When classifying fake GossipCop and ReCOVery articles, the models benefit from the multimodal learning algorithm the most when users are (partly) represented by their tweets. For example, the DistilBERT model returns the highest GossipCop results with the $+u/d+t$ setup ( -1.21/+4.68/+2.87\%; P/R/F1), while the CNN model achieves its highest ReCOVery prediction performance with the $+u/t$ setup (+3.19/+3.01/+3.12\%; P/R/F1). The multimodal learning algorithm has a rather limited positive impact on the HAN prediction performance across all three datasets. 

For the true class (Table \ref{tab:overall_results_true}), we observe similar user setup preferences as for the fake class. For the true Politifact articles, the models mainly benefit from the user setup with only profile descriptions ($+u/d$): +3.66/(+0.00)/+2.90\% (P/R/F1) for the CNN model, and +0.19/+8.00/+3.14\% (P/R/F1) for the DistilBERT model. When predicting true GossipCop and ReCOVery articles, the user setups that represent each user by their tweets (and profile description) yield the highest performance results. For example, DistilBERT's performance increases by +1.10/(-0.78)/+0.24\% (P/R/F1) for GossipCop articles with the $+u/d+t$ setup. For ReCOVery articles, CNN performance rises by +3.19/3.01/3.12\% (P/R/F1) with the $+u/t$ setup while HAN performance modestly increases by +0.15/1.50/0.85\% (P/R/F1) with the $+u/d+t$ setup.

\begin{table}
    \setlength{\tabcolsep}{1.2pt}
    \centering
    \small
    \begin{tabular}{c|c|ccc|ccc|ccc}
        \toprule
        && \multicolumn{3}{c}{Politifact} & \multicolumn{3}{c}{GossipCop} & \multicolumn{3}{c}{ReCOVery} \\ 
        \midrule
        && P & R & F1 & P & R & F1 & P & R & F1\\
        \midrule
        \parbox[t]{4mm}{\multirow{4}{*}{\rotatebox[origin=c]{90}{CNN}}} & \multirow{1}{*}{\textit{base}} & .7857 & .3333 & .4681 & \textbf{.7951} & .5542 & .6531 & .7872 & .5606 & .6549 \\
        \cmidrule{2-11}
        & \multirow{1}{*}{+u/d} & \textbf{.8235} & \textbf{.4242} & \textbf{.5600} & .7882 & \underline{.5591} & \textbf{.6542} & \underline{.7959} & \underline{.5909} & \underline{.6783} \\
        & \multirow{1}{*}{+u/t} & .7778 & \textbf{.4242} & \underline{.5490} & .7633 & \textbf{.5640} & .6487 & \textbf{.8750} & \textbf{.6364} & \textbf{.7368} \\
        & \multirow{1}{*}{+u/d+t} & .7647 & \underline{.3939} & \underline{.5200} & .7813 & .5542 & .6484 & \underline{.8478} & \underline{.5909} & \underline{.6964} \\
        \midrule
        \midrule
        \parbox[t]{4mm}{\multirow{4}{*}{\rotatebox[origin=c]{90}{HAN}}} & \multirow{1}{*}{\textit{base}} & \textbf{.8214} & \textbf{.6970} & \textbf{.7541} & .6684 & .6502 & .6592 & .7500 & \textbf{.8182} & .7826 \\
        \cmidrule{2-11}
        & \multirow{1}{*}{+u/d} & .7778 & .6364 & .7000 & \textbf{.6855} & .6281 & .6555 & \underline{.7571} & .8030 & .7794 \\
        & \multirow{1}{*}{+u/t} & .7778 & .6364 & .7000 & \textbf{.6855} & .6281 & .6555 & \underline{.7571} & .8030 & .7794 \\
        & \multirow{1}{*}{+u/d+t} & .7857 & .6667 & .7213 & .6569 & \textbf{.6650} & \textbf{.6610} & \textbf{.7714} & \textbf{.8182} & \textbf{.7941} \\
        \midrule
        \midrule
        \parbox[t]{4mm}{\multirow{4}{*}{\rotatebox[origin=c]{90}{DistilBERT}}} & \multirow{1}{*}{\textit{base}} & .8148 & \textbf{.6667} & .7333 & .7845 & .5468 & .6444 & .7463 & \textbf{.7576} & .7519 \\
        \cmidrule{2-11}
        & \multirow{1}{*}{+u/d} & \textbf{.8750} & .6364 & \textbf{.7368} & \textbf{.8045} & .5271 & .6369 & .7347 & .5455 & .6261 \\
        & \multirow{1}{*}{+u/t} & .7619 & .4848 & .5926 & .7786 & .5025 & .6108 & .7000 & .6364 & .6667 \\
        & \multirow{1}{*}{+u/d+t} & .8000 & .6061 & .6897 & .7724 & \textbf{.5936} & \textbf{.6731} & \textbf{.8421} & .7273 & \textbf{.7805} \\
        \bottomrule
    \end{tabular}
    \caption{Performance results for the \underline{fake} class: precision (P), recall (R) and F1-score (F1). The \underline{underlined} results indicate that the user-constrained model outperforms the corresponding $base$ model on that performance metric. The highest results for each model architecture (CNN, HAN, DistilBERT) are marked in \textbf{bold}.}
    \label{tab:overall_results_fake}
\end{table}

\begin{table}
    \setlength{\tabcolsep}{1.2pt}
    \centering
    \small
    \begin{tabular}{c|c|ccc|ccc|ccc}
        \toprule
        && \multicolumn{3}{c}{Politifact} & \multicolumn{3}{c}{GossipCop} & \multicolumn{3}{c}{ReCOVery} \\ 
        \midrule
        && P & R & F1 & P & R & F1 & P & R & F1\\
        \midrule
        \parbox[t]{4mm}{\multirow{4}{*}{\rotatebox[origin=c]{90}{CNN}}} & \multirow{1}{*}{\textit{base}} & .5000 & \textbf{.8800} & .6377 & .8722 & \textbf{.9552} & \textbf{.9118} & .8092 & .9248 & .8632 \\
        \cmidrule{2-11}
        & \multirow{1}{*}{+u/d} & \textbf{.5366} & \textbf{.8800} & \textbf{.6667} & \underline{.8732} & .9529 & .9113 & \underline{.8200} & .9248 & \underline{.8693} \\
        & \multirow{1}{*}{+u/t} & \underline{.5250} & .8400 & \underline{.6462} & \textbf{.8736} & .9451 & .9079 & \textbf{.8411} & \textbf{.9549} & \textbf{.8944} \\
        & \multirow{1}{*}{+u/d+t} & \underline{.5122} & .8400 & .6364 & .8718 & .9513 & .9098 & \underline{.8235} & \underline{.9474} & \underline{.8811} \\
        \midrule
        \midrule
        \parbox[t]{4mm}{\multirow{4}{*}{\rotatebox[origin=c]{90}{HAN}}} & \multirow{1}{*}{\textit{base}} & \textbf{.6667} & \textbf{.8000} & \textbf{.7273} & .8912 & .8988 & .8950 & .9055 & .8647 & .8846 \\
        \cmidrule{2-11}
        & \multirow{1}{*}{+u/d} & .6129 & .7600 & .6786 & .8863 & \textbf{.9096} & \textbf{.8978} & .8992 & \underline{.8722} & .8855 \\
        & \multirow{1}{*}{+u/t} & .6129 & .7600 & .6786 & .8863 & \textbf{.9096} & \textbf{.8978} & .8992 & \underline{.8722} & .8855 \\
        & \multirow{1}{*}{+u/d+t} & .6333 & .7600 & .6909 & \textbf{.8945} & .8910 & .8928 & \textbf{.9070} & \textbf{.8797} & \textbf{.8931} \\
        \midrule
        \midrule
        \parbox[t]{4mm}{\multirow{4}{*}{\rotatebox[origin=c]{90}{DistilBERT}}} & \multirow{1}{*}{\textit{base}} & .6452 & .8000 & .7143 & .8701 & .9529 & .9096 & \textbf{.8788} & .8722 & .8755 \\
        \cmidrule{2-11}
        & \multirow{1}{*}{+u/d} & \textbf{.6471} & \textbf{.8800} & \textbf{.7457} & .8661 & \textbf{.9598} & \underline{.9106} & .8000 & \underline{.9023} & .8481 \\
        & \multirow{1}{*}{+u/t} & .5405 & .8000 & .6452 & .8595 & \underline{.9552} & .9048 & .8273 & .8647 & .8456 \\
        & \multirow{1}{*}{+u/d+t} & .6061 & .8000 & .6897 & \textbf{.8811} & .9451 & \textbf{.9120} & .8732 & \textbf{.9323} & \textbf{.9018} \\
        \bottomrule
    \end{tabular}
    \caption{Performance results for the \underline{true} class: precision (P), recall (R) and F1-score (F1). The \underline{underlined} results indicate that the user-constrained model outperforms the corresponding $base$ model on that performance metric. The highest performance results for each model architecture (CNN, HAN, DistilBERT) are marked in \textbf{bold}.}
    \label{tab:overall_results_true}
\end{table}


%% file: discussion.tex
\section{Discussion}

We start with a general discussion of the performance results and continue to investigate the following research questions:
\begin{enumerate}
    \item \textit{Does leveraging user information lead to better latent article representations?}
    \item \textit{To which extent does user and tweet selection influence model performance?}
    \item \textit{Does the model indeed find and leverage user-article and user-user correlations?}
\end{enumerate}    
 
\subsection{General}
\begin{figure}
    \centering
    \includegraphics[width=8cm]{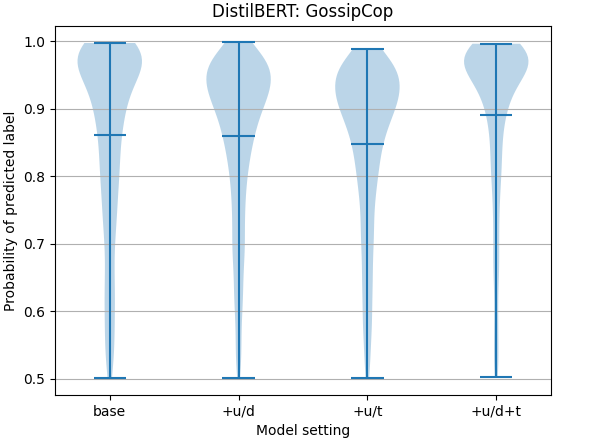}
    \caption{Distribution of the predicted probabilities yielded by the DistilBERT model for GossipCop articles in the test set. The blue horizontal lines indicate the minimum, mean and maximum probability value in the distribution.}
    \label{fig:violin}
\end{figure}

In terms of the performance metrics, the impact of the multimodal learning algorithm appears to be greater for the smaller datasets (Politifact and ReCOVery) than for the largest dataset (GossipCop). As the models are trained on all three datasets at the same time, the base model might have learned to discriminate between true and fake gossip news better than between more political or COVID-related news. This seems to suggest that leveraging additional user information adds little value to the prediction performance for GossipCop articles. Nonetheless, the multimodal learning algorithm might have influenced the models' confidence about their predictions. We therefore evaluate whether the models yield significantly higher prediction probabilities for GossipCop articles when leveraging user information. For each GossipCop article in the test set, we take the computed probability for the predicted label yielded by the model's softmax layer. This probability measure lies between 0.5 and 1, as the model outputs the label with highest probability. We then perform a two-sample T-test and Kruskal-Wallis H-test between the GossipCop prediction probabilities yielded by the $base$ model and those yielded by the model optimized using one of the three user setups ($+u/d$, $+u/t$, $+u/d+t$). 
Figure \ref{fig:violin} shows the distribution of the prediction probabilities yielded by each DistilBERT model. Both statistical tests confirm that the prediction confidence significantly differs between DistilBERT$_{base}$ ($\mu$ = 0.861697, $\sigma^2$ = 0.018571) and DistilBERT$_{+u/t}$ ($\mu$ = 0.847578, $\sigma^2$ = 0.014585) [T = 3.196966, $p$ < 0.01; H = 64.073104, $p$ < 0.01], and DistilBERT$_{base}$ and DistilBERT$_{+u/d+t}$ ($\mu$ = 0.891369, $\sigma^2$ = 0.015190)[T = -6.658362, $p$ < 0.01; H = 33.668810, $p$ < 0.01]. Only the Kruskal-Wallis H-test rejects the null hypothesis for DistilBERT$_{base}$ and DistilBERT$_{+u/d}$ ($\mu$ = 0.859106, $\sigma^2$ = 0.014207) [H = 23.783539, $p$ < 0.01]. The distributions depicted in Figure \ref{fig:violin} show that DistilBERT$_{+u/d+t}$ yields the statistically highest probabilities on average. The statistical tests confirm the same for the HAN model, but are inconclusive for the CNN model.

\subsection{Influence on Article Latent Space (RQ1)}
\label{monitoring:sec}

\begin{figure}
     \centering
     \begin{subfigure}[b]{0.225\textwidth}
         \centering
         \includegraphics[bb = 0 0 400 270 , clip = true, width=\textwidth]{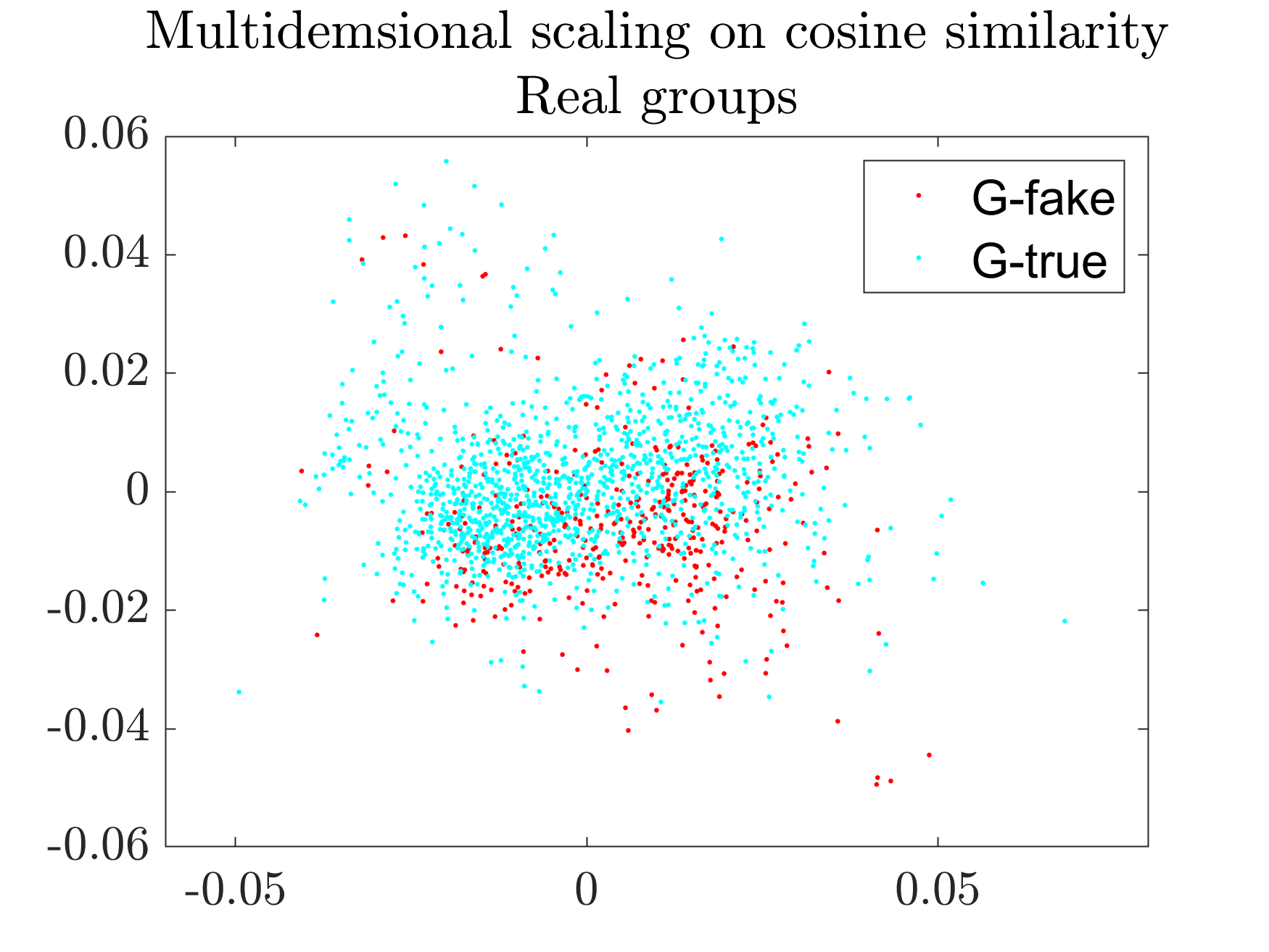}
         \caption{CNN$_{base}$ ($\bar{\omega}=0.5592$)}
         \label{fig:CNN_base}
     \end{subfigure}
     \hfill
     \begin{subfigure}[b]{0.225\textwidth}
         \centering
         \includegraphics[bb = 0 0 400 270 , clip = true, width=\textwidth]{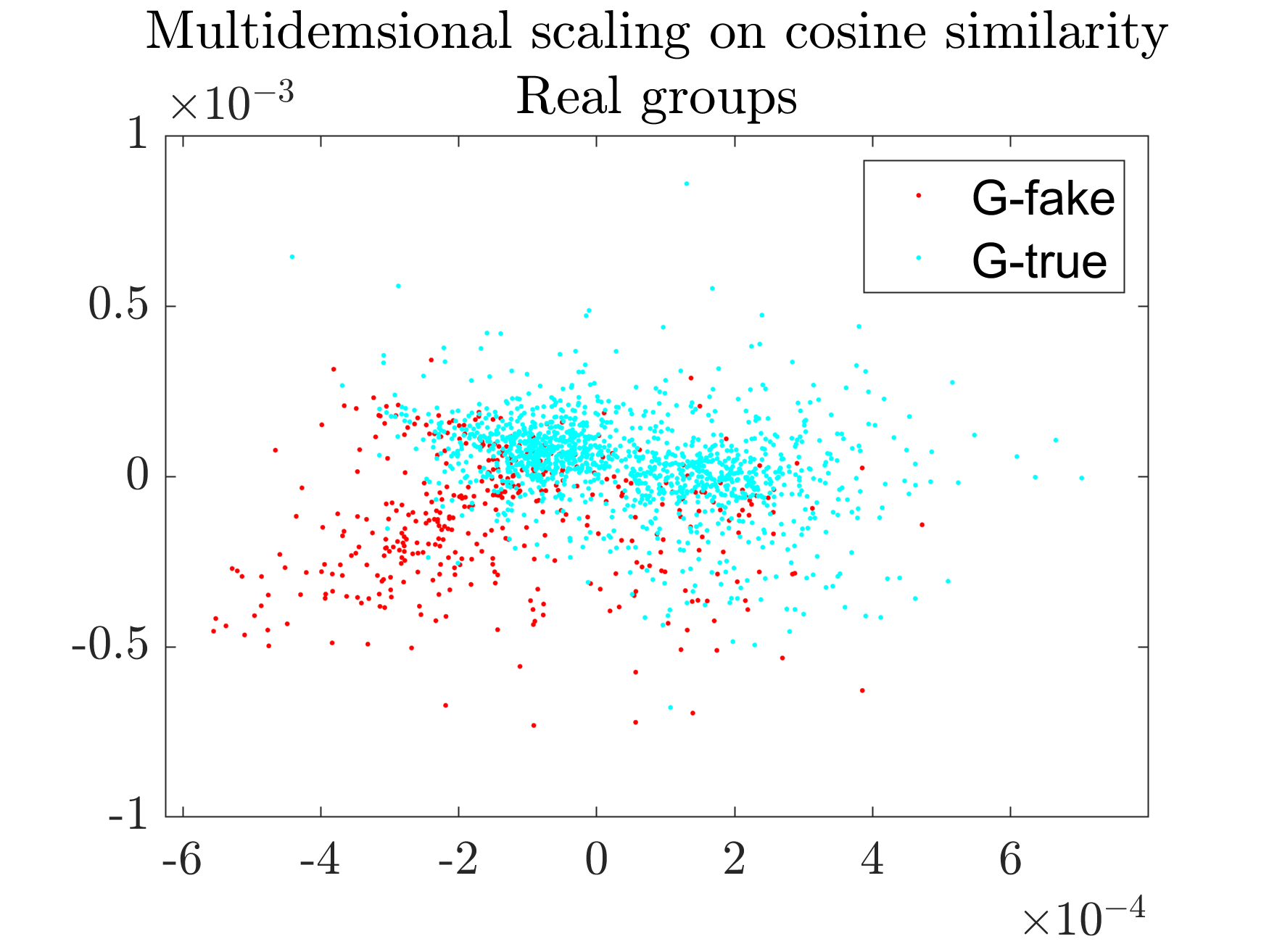}
         \caption{CNN$_{+u/d}$ ($\bar{\omega}=0.3897$)}
         \label{fig:CNN_user}
     \end{subfigure}\\[4mm]
     \begin{subfigure}[b]{0.225\textwidth}
         \centering
         \includegraphics[bb = 0 0 400 270 , clip = true, width=\textwidth]{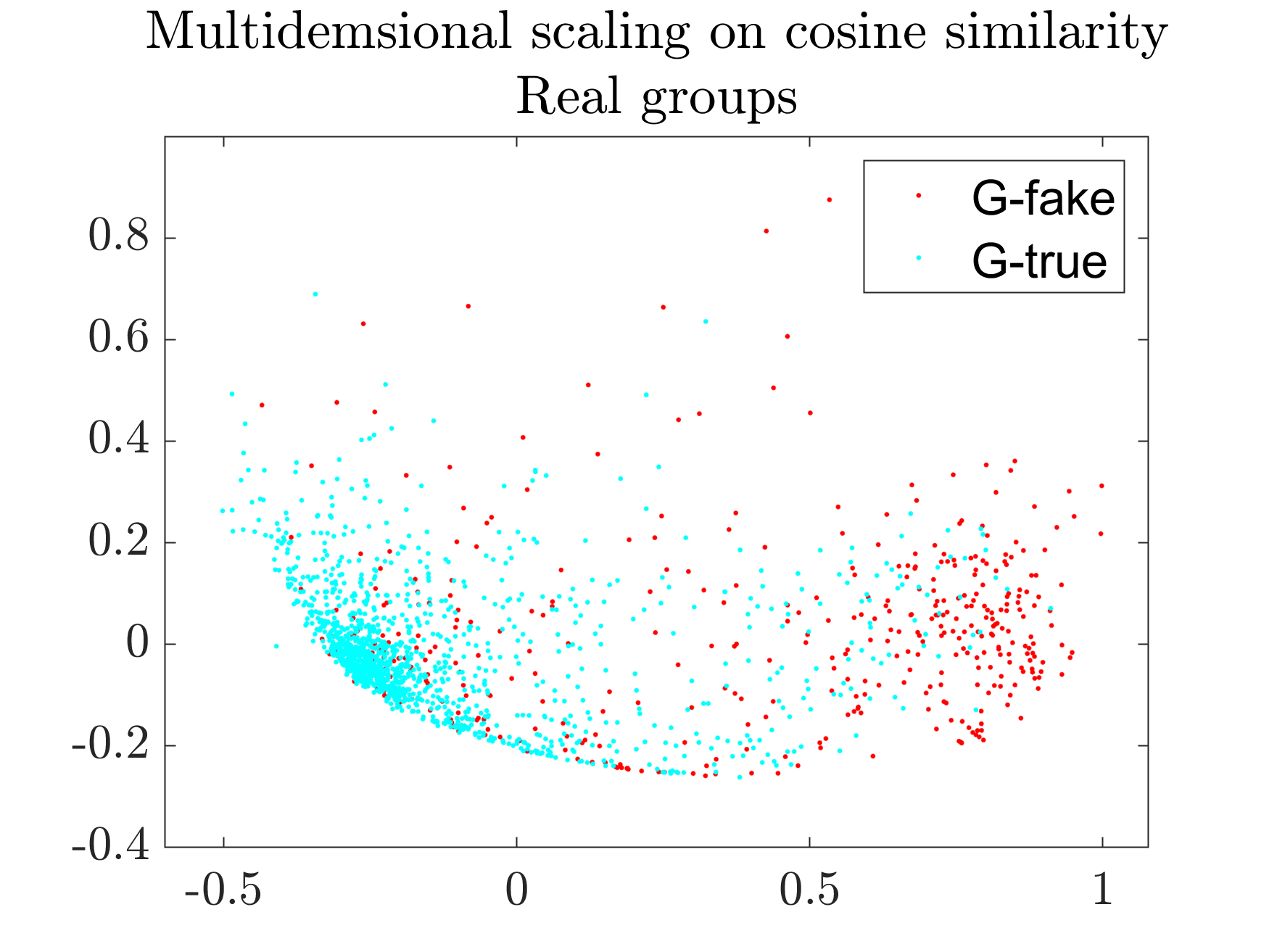}
         \caption{HAN$_{base}$ ($\bar{\omega}=0.2635$)}
         \label{fig:HAN_base}
     \end{subfigure}
    \hfill
     \begin{subfigure}[b]{0.225\textwidth}
         \centering
         \includegraphics[bb = 0 0 400 270 , clip = true, width=\textwidth]{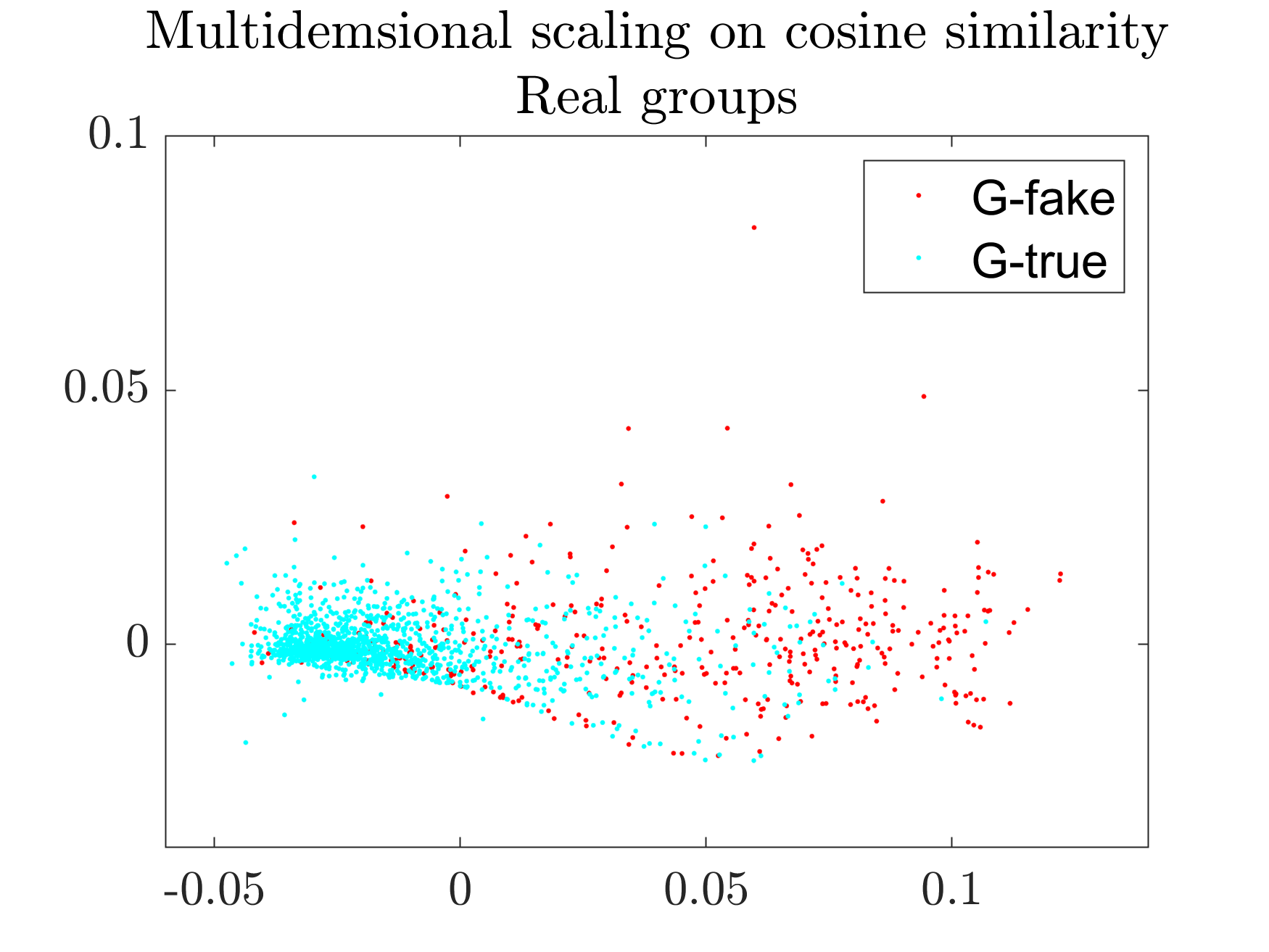}
         \caption{HAN$_{+u/d}$ ($\bar{\omega}=0.2081$)}
         \label{fig:HAN_user}
     \end{subfigure}\\[4mm]
     \begin{subfigure}[b]{0.225\textwidth}
         \centering 
         \includegraphics[bb = 0 0 400 270 , clip = true,  width=\textwidth]{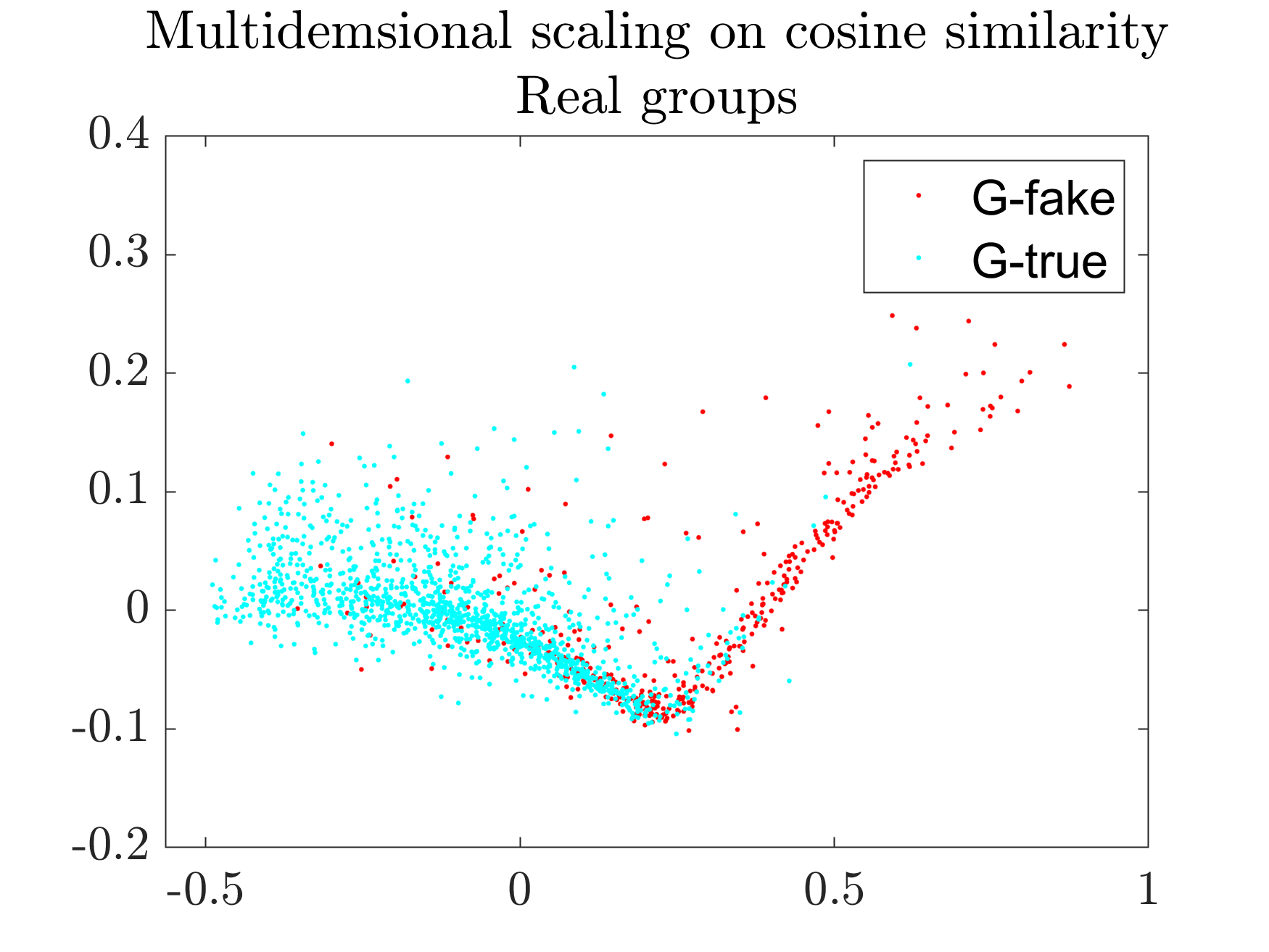}
         \caption{DistilBERT$_{base}$ ($\bar{\omega}=0.0432$)}
         \label{fig:BERT_base}
     \end{subfigure}
     \hfill
     \begin{subfigure}[b]{0.225\textwidth}
         \centering
         \includegraphics[bb = 0 0 400 270 , clip = true, width=\textwidth]{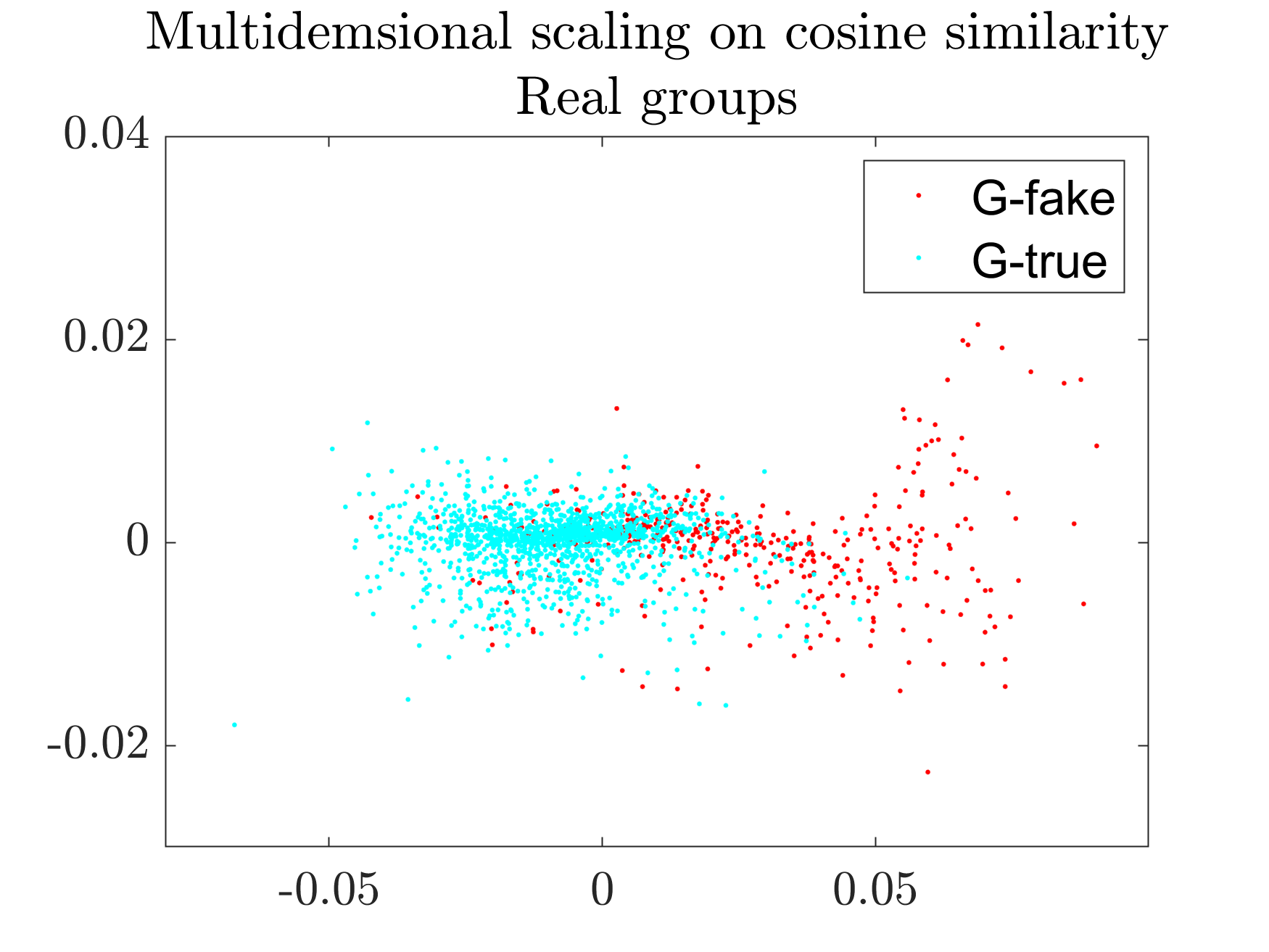}
         \caption{DistilBERT$_{+u/d}$ ($\bar{\omega}=0.0807$)}
         \label{fig:BERT_user}
     \end{subfigure}
        \caption{Statistical analysis: multidimensional scaling based on cosine similarity for GossipCop latent article representations (test set). The average overlap  between the two groups is quantified by $\bar{\omega}$.}
        \label{fig:MULTI_SCALING}
\end{figure}

The multimodal learning algorithm constrains the classification parameters by enforcing a minimized cosine distance between latent article and user representations and a minimized cosine distance between latent user representations during training. We analyze how and to which extent these user-related constraints influence the article latent space.

The effect of the latent space choice in our experimental settings can be appreciated using some established dimension reduction techniques. We illustrate here two possible approaches: multidimensional scaling \cite{CoxCox:1994} and principle component analysis (PCA) \cite{HubRouBra:2005,HubRouVer:2009}. In both cases, we applied a robust variant of the methodologies to avoid that results are distorted by the presence of outliers and deviations from canonical distributional assumptions. 
In the case of multidimensional scaling, the panels of Figure \ref{fig:MULTI_SCALING} show a two-dimensional representation of the $n$ rows of the latent space, where the Euclidean distances between them approximate a monotonic transformation of the corresponding $n \times n$ cosine similarities computed in the original space. Robustness is ensured by applying Huber's weighting in the estimation, using MATLAB function \texttt{mdscale} with option \texttt{statset('Robust','on',   'RobustWgtFun','huber')}. 
\begin{figure}
    \centering
    \begin{subfigure}[b]{0.23\textwidth}
         \centering
         \includegraphics[bb = 0 0 304 270 , clip = true, width=\textwidth]{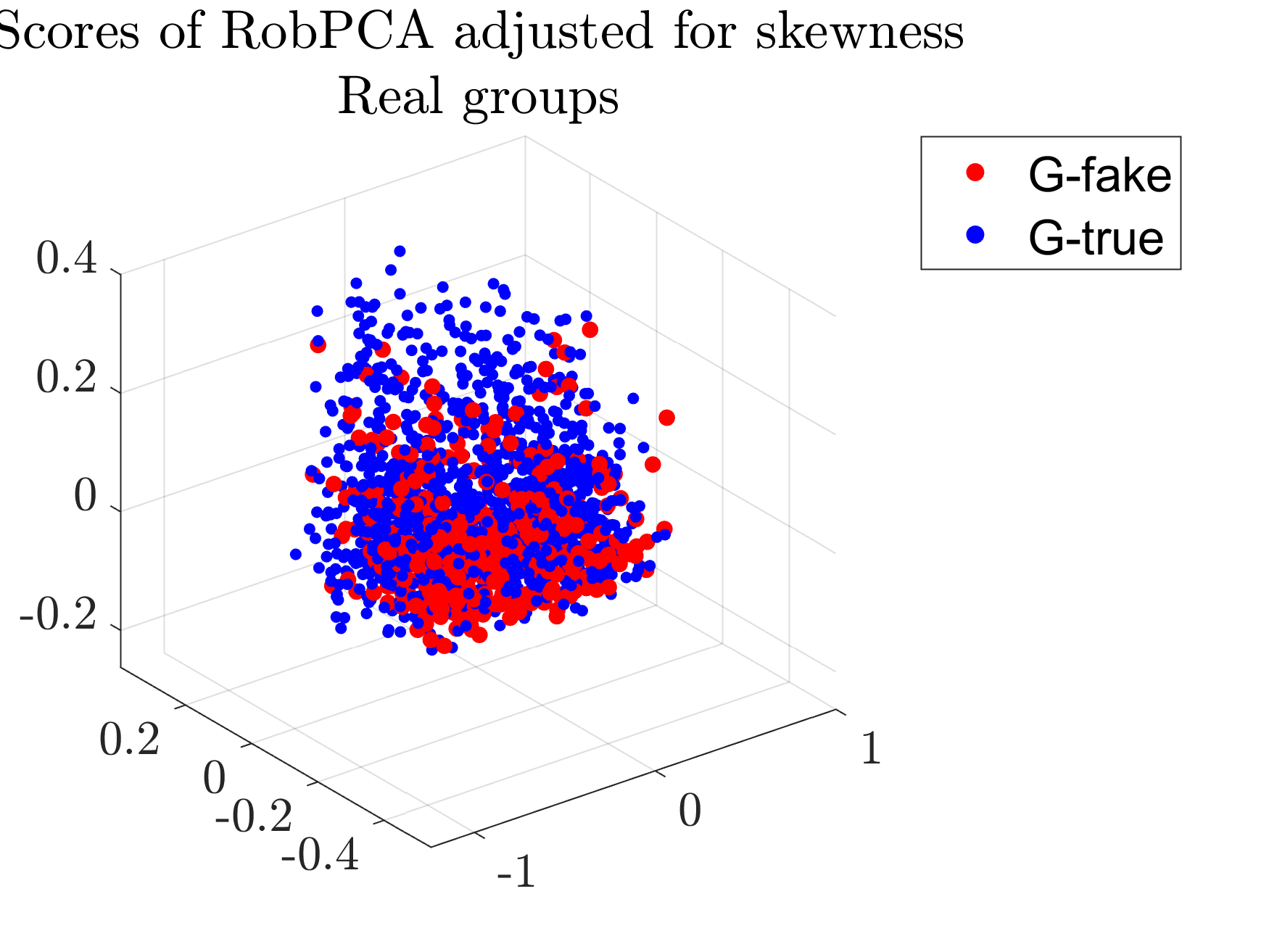}
         \caption{CNN: base ($\bar{\omega}=0.7700$)}
         \label{fig:y equals x}
     \end{subfigure}
     \hfill
     \begin{subfigure}[b]{0.23\textwidth}
         \centering
         \includegraphics[bb = 0 0 304 270 , clip = true, width=\textwidth]{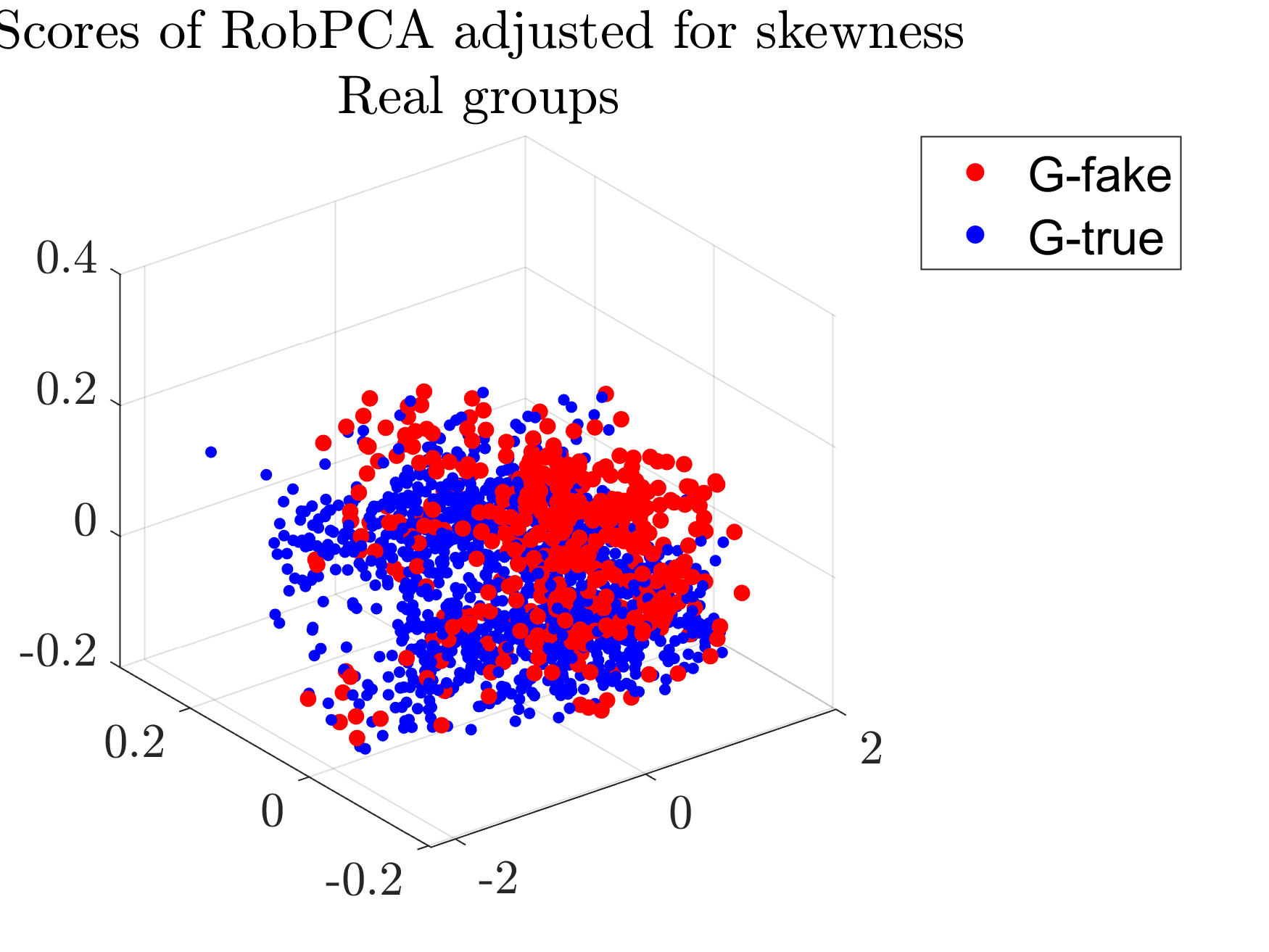}
         \caption{CNN: +u/d ($\bar{\omega}=0.3966$)}
         \label{fig:three sin x}
     \end{subfigure}
    \caption{Statistical analysis: scores of the three main principle components. Validation set, GossipCop. The average overlap  between the two groups is quantified by $\bar{\omega}$.}
    \label{fig:robPCA}
\end{figure}
The robust PCA is applied using function \texttt{robpca} of the LIBRA toolbox \cite{libra}. We also used additional monitoring features of the FSDA toolbox that allow removing an appropriate percentage of outliers from the analysis and study the fine-grained structure of the data (number of groups) \cite{RiaPerTor:2012, TorRiaMor:2021}.
\footnote{The MATLAB code used for this part of the data analysis is available upon request.} 

Figures  \ref{fig:MULTI_SCALING} and \ref{fig:robPCA} show how the true (in blue) and fake (in red) classes are projected in the two sub-space representations. It is clear from these examples that implementing our multimodal learning algorithm increases the separation between the two prediction classes in the article latent space. \citet{MM:2010} provide a possible approach to quantify this separation, with an overlap measure that expresses the probability of miss-classification assuming a Gaussian mixture model as data generating process. We computed this measure using the FSDA \texttt{overlap} function described in \citet{RCPT:2015}, which is reported in the caption of the figures with $\bar{\omega}$. Note that the measure is quite reliable in the PC representations of Figure \ref{fig:robPCA}, where the Gaussian mixture is clearly appropriate, while in some cases the multidimensional scaling (Figure \ref{fig:MULTI_SCALING}) produces groups that are quite skewed (e.g., panel \ref{fig:BERT_base}) and this might bias the proposed separation estimate. 
In sum, the statistical visualization techniques and overlap measures show that our multimodal learning algorithm forces the classification models to learn feature representations that better discriminate between fake and true news texts. 

\subsection{User and Tweet Selection (RQ2)}

We assess to which extent the selection of both users and tweets in the user subsets influence model performance.

\begin{figure}
    \centering
    \includegraphics[width=8.5cm]{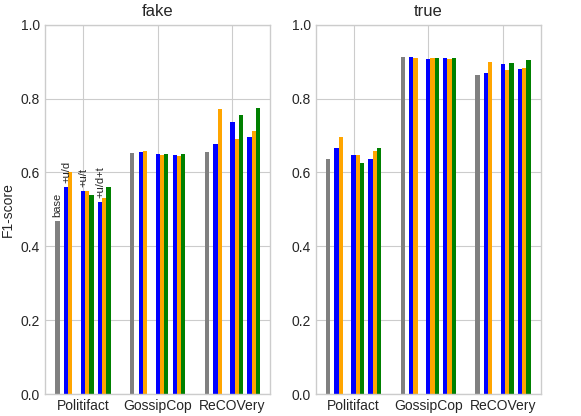}
    \caption{Impact of altered user (in orange) and tweet (in green) selection during training on the F1-score for the fake (left figure) and true (right figure) class in the test set: CNN$_{base}$ (in grey), CNN$_{+u/d}$, CNN$_{+u/t}$ and CNN$_{+u/d+t}$ (in blue).}
    \label{fig:user_tweet_selection}
\end{figure}

We start with the \textbf{user selection}. In the above experiments (Section \ref{experiments}), the $S$ (= 10) user IDs linked to the lowest tweet IDs in an article's tweet ID list are automatically included in the article's user subset. This way, each user subset contains users who were among the firsts to share the article on Twitter. In other words, the user subset reflects the audience of the article's \textit{early dissemination process}. We investigate to which extent model performance changes when using users from the articles' \textit{late dissemination process}. For this, we select the user IDs linked to the highest tweet IDs in an article's tweet ID list and train the models with the new user subsets. Model performance is conjectured to be higher with the original, early dissemination subset than with the new, late dissemination subset. We expect that the correlation between an article and the early dissemination users is higher, as it is more likely that the article's author and news outlet are among the first to spread their articles on social media. For brevity, we only show and discuss the results for the CNN model (Figure \ref{fig:user_tweet_selection}; orange bars). 

The late dissemination user subset raises model performance for the Politifact and ReCOVery dataset when using their description (CNN$_{+u/d}$) and both description and tweets (CNN$_{+u/d+t}$). Model performance remains more or less the same for the GossipCop dataset with all user setups. The results with the new user subsets thus rejects our hypothesis that leveraging early dissemination users leads to higher performance increases.

Not only the user selection, but also the \textbf{tweet selection} might influence model performance. Currently, the most recent tweets in a user's timeline are taken into consideration in the $+u/t$ and $+u/d+t$ setups. Upon investigating the time span of user timeline of 10,000 randomly selected users from the user subsets, we found that more than half of the timelines span max. three months, and less than one in four have a time span of more than a year. By taking the \textit{oldest tweets} from a user's timeline instead of the \textit{newest tweets}, the user representation will reflect a slightly older user identity that may correlate more with the earlier spread articles. Given the articles' publication date in the datasets (i.e., Politifact: 2008-2018, GossipCop: 2017-2018, reCOVery: 2020), we expect that the models' prediction performance for reCOVery articles will increase as the oldest tweets in the user timelines might cover more diverse COVID-related topics. For the other two datasets, on the other hand, performance should not be affected too much. The results for the CNN model (Figure \ref{fig:user_tweet_selection}; green bars) confirm our hypotheses: the model performs consistently better for ReCOVery articles, especially for the fake class, while its performance for the other two datasets remains rather stable. This suggests that leveraging user information from the same time period as the article that is evaluated improves model performance. 

\subsection{Correlations: Experimental Analysis (RQ3)}

When implementing the multimodal learning algorithm, does the model learn more discriminative article features because there are indeed correlations between the user-generated content and the news article content, and between the user-generated content of the users who shared the same article, or simply because there is more data? To investigate this, we first pair each article with another, randomly chosen user subset (random seed = 42). In this first experiment (\textit{Random Subset}), the users contained within the same user subset remain unchanged, but they are linked to another news article. This way, we distort possible user-article correlations while maintaining possible correlations between users in the same user subset. We expect that model performance will drop below base model performance, as incorrect user-article correlations will be learned. In a second experiment (\textit{Random Subset + Composition}), we take it a step further and also change the composition of the user subsets by randomly assigning a user to a different user subset (random seed = 42). We expect that this experimental setup will yield the lowest performance results as both the article-user and user-user correlations are distorted. 

\begin{figure}
    \centering
    \includegraphics[width=8.5cm]{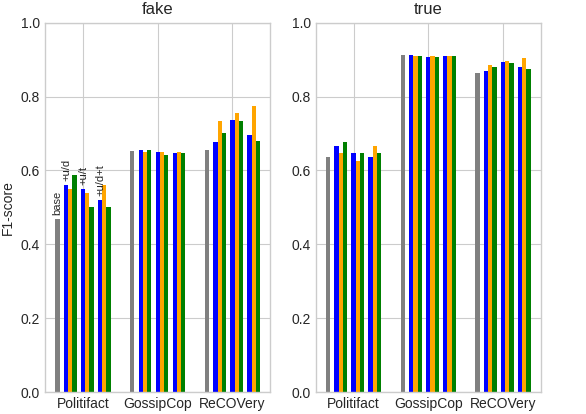}
    \caption{Impact of distorting the user-article correlations (\textit{Random Subset}; in orange), and distorting both user-article and user-user correlations (\textit{Random Subset + Composition}; in green) during training on the F1-score for the fake class (left figure) and true class (right figure) in the test set: CNN$_{base}$ (in grey), CNN$_{+u/d}$, CNN$_{+u/t}$ and CNN$_{+u/d+t}$ (in blue).}
    \label{fig:experimental_correlation}
\end{figure}

Figure \ref{fig:experimental_correlation} reports the F1-score for the CNN models for the \textit{Random Subset} experiment (orange bars) and the \textit{Random Subset + Composition} experiment (green bars). Overall, the \textit{Random Subset} experiment yields counter-intuitively higher results than the original user-constrained CNN models (CNN$_{+u/d}$, CNN$_{+u/t}$, CNN$_{+u/d+t}$). Although the \textit{Random Subset + Composition} experiment lowers the performance of those user-constrained models, they still perform better than the CNN$_{base}$ model. It can thus be argued that the user-constrained models benefit from leveraging extra user data by having distance constraints between and within the two modalities enforced during model optimization without actually uncovering the assumed real-world correlations constructed by the user modality. 

\subsection{Correlations: Qualitative Analysis (RQ3)}

\begin{figure}[]
    \centering
    \includegraphics[width=8.5cm]{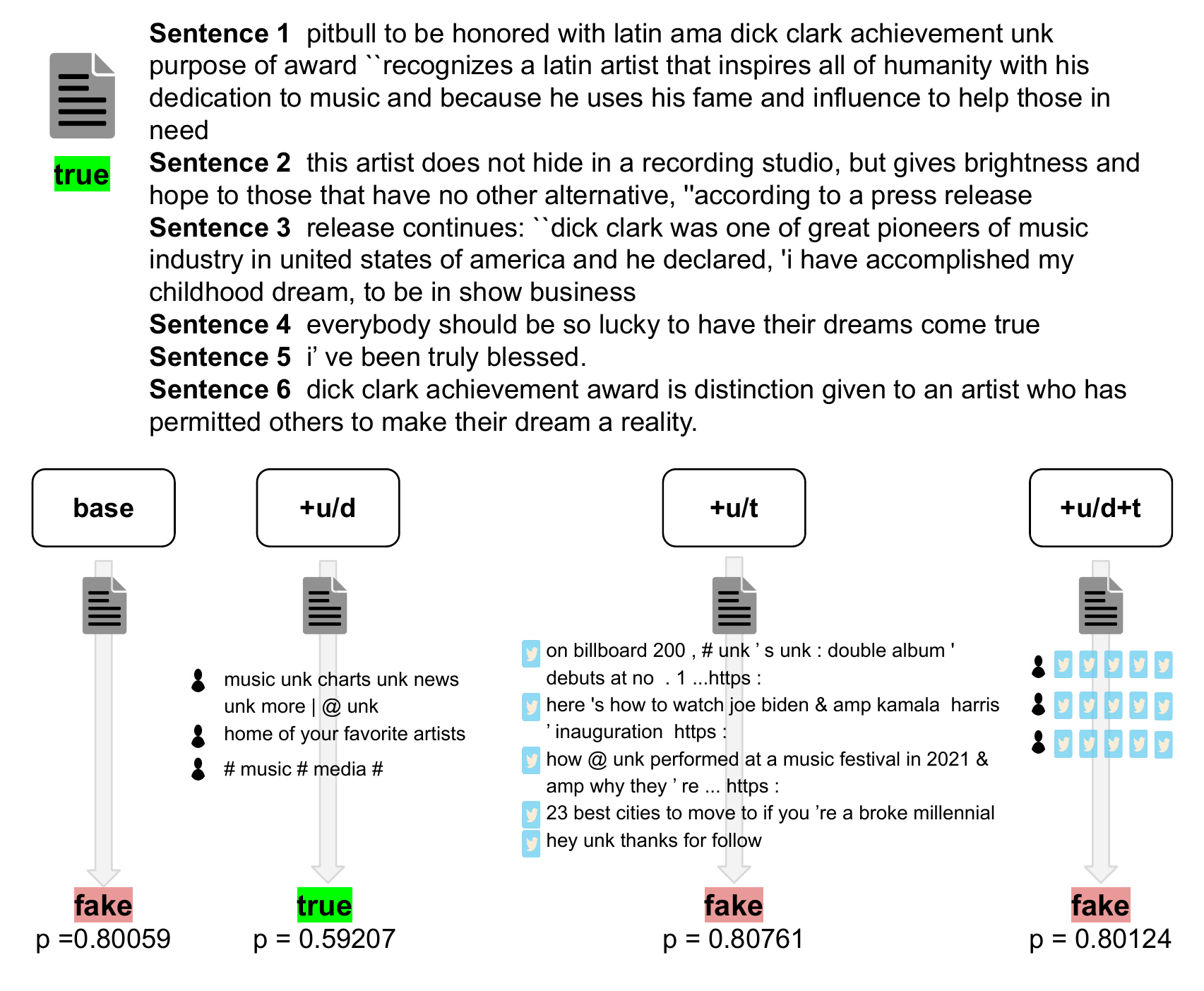}
    \caption{Example from the GossipCop dataset (training set) with output from the HAN models: HAN$_{base}$, HAN$_{+u/d}$, HAN$_{+u/t}$ and HAN$_{+u/d+t}$.}
    \label{fig:qualitative_correlations}
\end{figure}

In this brief qualitative analysis, we try to check whether we as humans can detect correlations between the news article text and the user-generated content (i.e., profile description and tweets) that possibly guide a model towards the correct prediction label. We randomly select 100 samples from the training set (random seed = 42) with the predicted label yielded by the HAN models and focus on cases where the models do not agree on the label. Figure \ref{fig:qualitative_correlations} provides such a case. In that example, the HAN$_{+u/d}$ predicts the ground-truth true label whereas the other models assign similar high probabilities to the incorrect fake label. It could be argued that the presence of similar topics in the news article and profile descriptions (i.e., music) guides the model towards the correct prediction label. Those topics are also found in the tweet timelines, but both HAN$_{+u/t}$ and HAN$_{+u/d+t}$ predict the opposing label with a high probability nevertheless. The tweets discuss a wide variety of topics and, therefore, introduce additional noise in the user representations. That noise seems to eclipse the valuable information contained in the user descriptions in HAN$_{+u/d+t}$. We observe similar patterns in other training samples. Given that user-generated content is inherently noisy, future work can look into various filtering techniques to distinguish useful and relevant information from highly noisy and irrelevant Twitter content before leveraging it in a multimodal learning algorithm.   

\subsection{Limitations}

Fake news articles and creators tend to quickly disappear from the Internet \cite{allcott2017social}. As a result, not all fake news articles and profiles can be extracted, leading to a reduced share of fake news articles in the datasets and possibly incomplete Twitter dissemination processes.
The datasets will thus become obsolete in the course of time. 
Moreover, our approach relies on user-generated Twitter profile descriptions and tweets from user timelines. A user can alter their description and generate new tweets after sharing the article that is evaluated. As time passes by, a user's identity reflected in their current description and latest tweets might no longer align with their identity at the time of sharing the article. 

%% file: conclusion.tex
\section{Conclusion}

This paper introduced a novel, multimodal learning algorithm that allows a disinformation detection model to learn features that discriminate better between true and false news using user-generated content on Twitter. It provides an elegant way to integrate and correlate multimodal information without requiring its presence at testing time. Further research can build on this approach and experiment with other modalities such as images and audio. Furthermore, the statistically robust approach we proposed to check the effect of the user-related constraints on the article latent space has a wider potential: it can be applied to understand the fine-grained structure of the data used for training or validating models, and to clean them if necessary. As it can avoid introducing disturbances in the model estimates, this would arguably lead to better generalization performances.